\documentclass[11pt]{article}
\usepackage[utf8]{inputenc}
\usepackage[T1]{fontenc}
\usepackage[USenglish]{babel}
\usepackage{csquotes}
\usepackage{amsmath, amssymb, amsthm, amsfonts}

\usepackage{color}
\usepackage{placeins}

\usepackage{hyperref}
\usepackage{layouts}

\usepackage[a4paper, total={5.8in, 8.5in}, footskip=40pt, footnotesep=20pt]{geometry}
\usepackage{authblk}

\usepackage[ttscale=.85]{libertine}
\usepackage{libertinust1math}
\usepackage[activate={true,nocompatibility}, final, tracking=true, kerning=true, spacing=true, factor=1100, stretch=10, shrink=10]{microtype}
\microtypecontext{spacing=nonfrench}

\SetExtraSpacing{ encoding = {TS1} } { }

\usepackage{fnpct} 

\usepackage[
backend=bibtex,
style=numeric,
maxcitenames=2,
natbib
]
{biblatex}

\emergencystretch=\maxdimen
\usepackage{xcolor}
\hypersetup{
    colorlinks,
    linkcolor={red!50!black},
    citecolor={blue!50!black},
    urlcolor={blue!80!black}
}

\usepackage{multirow}

\begin{document}

\title{The Impact of Preprocessing Methods on Racial Encoding and Model Robustness in CXR Diagnosis}

\author[*]{Dishantkumar Sutariya}
\author[*$\dagger$]{Eike Petersen}
\affil[*]{Fraunhofer Institute for Digital Medicine MEVIS}
\affil[$\dagger$]{Corresponding author: \href{mailto:eike.petersen@mevis.fraunhofer.de}{\texttt{eike.petersen@mevis.fraunhofer.de}}}

\date{}

\maketitle

\begin{abstract}
Deep learning models can identify racial identity with high accuracy from chest X-ray (CXR) recordings.
Thus, there is widespread concern about the potential for racial shortcut learning, where a model inadvertently learns to systematically bias its diagnostic predictions as a function of racial identity.
Such racial biases threaten healthcare equity and model reliability, as models may systematically misdiagnose certain demographic groups.
Since racial shortcuts are diffuse -- non-localized and distributed throughout the whole CXR recording -- image preprocessing methods may influence racial shortcut learning, yet the potential of such methods for reducing biases remains underexplored.
Here, we investigate the effects of image preprocessing methods including lung masking, lung cropping, and Contrast Limited Adaptive Histogram Equalization (CLAHE). These approaches aim to suppress spurious cues encoding racial information while preserving diagnostic accuracy.
Our experiments reveal that simple bounding box-based lung cropping can be an effective strategy for reducing racial shortcut learning while maintaining diagnostic model performance, bypassing frequently postulated fairness-accuracy trade-offs.
\end{abstract}

\section{Introduction}
The integration of artificial intelligence (AI) into medical imaging holds transformative promise. 
However, the deployment of these systems in real-world healthcare settings has revealed critical concerns about their fairness and reliability across diverse patient populations. One alarming finding concerns the potential for racial bias in AI models for chest X-ray (CXR) disease diagnosis~\cite{seyyed2021underdiagnosis,gichoya2022ai,burns2023ability}. These models have been shown to be capable of inferring race from standard CXR recordings with very high accuracy~\cite{gichoya2022ai} -- despite race being imperceptible to human radiologists.
This raises the potential for racial shortcut learning, where a model trained for disease classification might inadvertently exploit racial correlations for disease classification~\cite{gichoya2022ai,Glocker2023,Glocker2023a}.
While it is generally challenging to assess whether a potential racial shortcut is indeed exploited~\cite{Glocker2023,Glocker2023a}, such disparities threaten to undermine the trust and safety of AI in clinical practice~\cite{seyyed2021underdiagnosis}.

Racial shortcuts are particularly challenging to assess and mitigate because of their diffuse nature:  
racial prediction does not rely on localized features but rather depends on image features distributed throughout the whole recording~\cite{gichoya2022ai}, with race classification feasible even based on just the grayscale histogram of a recording~\cite{burns2023ability}.
Image preprocessing techniques may therefore have an effect on the potential for demographic (racial) shortcut learning, but the extent to which such methods can help reduce biases 
remains underexplored.

To address this gap, we here investigate the potential of three generic CXR preprocessing methods to alleviate biases while maintaining diagnostic performance.
We implement and evaluate two lung masking strategies as well as Contrast-Limited Adaptive Histogram Equalization (CLAHE) to restrict and enhance model focus to clinically relevant regions, thereby potentially suppressing demographic confounders embedded in the imaging data. 
We evaluate (internally and externally) racial encoding as well as diagnostic performance in models trained and evaluated on such preprocessed recordings.
Among other findings, we show that simple bound box-based lung cropping can effectively reduce racial encoding while maintaining overall diagnostic accuracy.

\section{Materials and methods}

\subsection{Related work}
Gichoya et al.~\cite{gichoya2022ai} demonstrated that AI models can infer patients' racial or ethnic identity from chest radiographs with high accuracy, even when clinicians cannot. This ability to predict self-reported race with high AUROC scores across multiple modalities, imaging vendors, and clinical tasks is concerning, as it suggests that AI models may be relying on non-clinical, demographically linked signals rather than focusing on medically meaningful patterns ~\cite{burns2023ability,Glocker2023}.
Gichoya et al.~\cite{gichoya2022ai} already demonstrated that racial shortcuts are non-localized and diffuse, and Burns et al.~\cite{burns2023ability} showed that racial identity can even be identified from grayscale intensity histograms alone, disregarding all structural information.
Technical recording parameters such as the view positioning have been found to be major contributors to race detection capability~\cite{lotter2024acquisition}, and Wang et al.~\cite{wang2024drop} showed that generic data augmentation can substantially reduce the potential for demographic shortcut learning in CXR diagnosis.

That models \emph{can} learn to identify racial identity does not necessarily imply that they \emph{do} exploit this potential shortcut. 
Glocker et al.~\cite{Glocker2023,Glocker2023a} investigated demographic encoding in CXR diagnosis models, observing small but statistically significant differences in latent embeddings between racial groups after accounting for various confounding factors.
These differences were more pronounced in a foundation model compared to a single-disease model~\cite{Glocker2023a}.
Still, racial `encoding' in diagnostic (single-disease) models was similar to an ImageNet-trained baseline model, indicating that racial shortcut learning occurred only to a minor degree, if at all.
Seyyed-Kalantari et al.~\cite{seyyed2021underdiagnosis} demonstrated that standard CXR diagnosis models exhibit systematic underdiagnosis for Black patients across multiple disease labels, but such disparities tend to disappear after controlling for confounding factors such as age, sex, and disease distribution~\cite{Glocker2023}.
Nevertheless, due to the gravity of concerns about potential racial biases, the need for generic methods that can reduce the potential for demographic biases while maintaining or potentially even improving diagnostic accuracy remains.

\subsection{Datasets}
\paragraph{MIMIC-CXR}
The MIMIC-CXR-JPG dataset~\cite{johnson2019mimic,Johnson2024} comprises 377,110 images across 227,835 studies, covering 65,379 patients. 
Following the findings of Lotter~\cite{lotter2024acquisition}, we exclude lateral views and retain only frontal (AP/PA) recordings. Following Weng et al.~\cite{Weng2023}, we discard multiple recordings for the same patient and keep just one out of the set with the most disease labels provided to minimize the risk of label errors. From the resulting 41,168 recordings, we construct a test set by randomly sampling 35 positive instances for each of the 11 disease labels for each of the top-4 racial/ethnic groups (White, Black, Asian, Hispanic), resulting in a total test-set size of 1,430 samples. The remaining data are randomly split into a training and validation set of 34,400 (95\%) and 1,811 (5\%) samples, respectively. We ensure that there is no patient overlap between any of the three sets.

\paragraph{CheXpert}
The CheXpert dataset~\cite{irvin2019chexpert} comprises 224,316 CXR recordings and diagnostic labels from 65,240 patients. We use the same sampling strategies (frontal views, one recording per patient) and racial/ethnic groups as in MIMIC-CXR.
The entire resulting set of 51,627 samples is used for external evaluation only.

\subsection{Experimental setup}
We finetune an ImageNet-pretrained DenseNet-121 for multi-label disease classification on MIMIC-CXR and evaluate diagnostic performance as well as racial encoding on the MIMIC-CXR test set and externally on CheXpert.
For all purposes, recordings were downscaled to $224 \times 224$.
For optimization, we use AdamW, label smoothing (0.1), batch size 8, initial learning rate (LR) $10^{-4}$, cosine annealing over 30 epochs (decreasing to a minimum LR of $10^{-6}$), and early stopping (patience 5) on validation AUROC.
We use standard image augmentations (random rotations by $\pm10^{\circ}$ and random horizontal / vertical flips).
Racial encoding is evaluated by freezing the image encoder and training a race classification head.
We rerun all experiments with 5 different seeds.
All experiments were implemented using PyTorch and executed on NVIDIA A100 GPUs.
Our full code is available at \url{https://github.com/dishant24/BVM_Chest_X-Ray_Fair_AI}.

\subsection{Preprocessing methods}
To reduce race-correlated pixel-level signals in CXR recordings and prevent models from learning demographic shortcuts, we investigate different preprocessing methods. The aim is to suppress racial encoding while preserving clinical features, thereby potentially improving fairness, robustness, and generalization across racial groups.

\paragraph{Contrast limited adaptive histogram equalization (CLAHE)}
Unlike standard histogram equalization that operates on the entire image, CLAHE~\cite{zuiderveld1994contrast} processes images in small, non-overlapping ($8 \times 8$) tiles. For each tile, it computes a local intensity histogram, clips it to a specified limit (to prevent over-amplification of noise, we use 2) and redistributes the clipped portion to ensure better contrast across the entire intensity range. The cumulative distribution function (CDF) is computed for each tile to create a local mapping function. The per-tile mappings are then recombined using bilinear interpolation, yielding the enhanced output image. CLAHE boosts local contrast without causing the over-amplification of noise often seen with global histogram equalization, emphasizing fine details and potentially aiding the model's ability to discern pathological patterns.

\paragraph{Lung masking}
To mitigate the influence of confounding factors and potentially enhance model fairness and generalizability by restricting model attention to clinically relevant image regions, we implement lung masking as a preprocessing step similar to the approaches taken by Aslani et al.~\cite{aslani2022optimising} and Sourget et al.~\cite{Sourget2025}.
We utilize lung segmentation masks from CheXmask~\cite{gaggion2024chexmask}, a high-quality lung segmentation mask dataset for CXR recordings.  The dataset includes reliability scores (RCA) for each segmentation, ensuring mask quality. 
We filtered samples to include only those with an average Dice RCA greater than 0.7. 
A morphological dilation with a 60-pixel margin (on the original mask resolution of $1024 \times 1024$) was applied to preserve diagnostically relevant contextual features near lung and heart boundaries. 

\paragraph{Lung cropping}
As an alternative to lung masking, we cropped the image to a bounding box containing the whole lung musk. This approach avoids the introduction of abrupt, high-intensity transitions at mask edges and reduces the risk of models exploiting artificial boundaries and intensity fluctuations as shortcuts.

\section{Results}
Table~\ref{tab:combined_auroc} presents the main results of our experiments.
Diagnostic performance on the internal (MIMIC) test set is consistent across preprocessing methods and comparable to the baseline model without preprocessing, with a small disadvantage for the lung masking approach.
On the external (CheXpert) evaluation set, this disadvantage is more pronounced (lung masking AUROC 0.696 vs. 0.742 for the baseline), with the other two methods (CLAHE and lung cropping) performing similarly to the baseline.

\begin{table}[t]
    \centering
    \caption{Comparison of preprocessing methods on racial and diagnostic AUROC, mean $\pm$ standard deviation across repetitions.}
    \label{tab:combined_auroc}
\begin{tabular*}{\textwidth}{@{\extracolsep{\fill}} l cc @{\hspace{1.2em}} cc @{}}
        \hline
        \noalign{\vskip 2pt}
        \multirow{2}{*}{Method} & \multicolumn{2}{c}{Race AUROC} & \multicolumn{2}{c}{Diagnostic AUROC} \\
        & Internal & External & Internal & External \\
        \noalign{\vskip 2pt}
        \hline
        \noalign{\vskip 2pt}
        Baseline & $0.639 \pm 0.003$ & $0.623 \pm 0.004$ & $0.764 \pm 0.003$ & $0.742 \pm 0.001$ \\
        Masking & $0.630 \pm 0.005$ & $0.566 \pm 0.013$ & $0.759 \pm 0.002$ & $0.696 \pm 0.005$ \\
        Cropping & $0.641 \pm 0.006$ & $0.593 \pm 0.007$ & $0.763 \pm 0.003$ & $0.738 \pm 0.004$ \\
        CLAHE & $0.642 \pm 0.007$ & $0.624 \pm 0.024$ & $0.765 \pm 0.004$ & $0.738 \pm 0.005$ \\
        \noalign{\vskip 2pt}
        \hline
    \end{tabular*}
\end{table}

In terms of racial encoding, all models enable race prediction with above-chance accuracy, as expected~\cite{gichoya2022ai,Glocker2023,Glocker2023a}.
On the internal (MIMIC) test set, all methods yield similar race identification performance.
Externally (on CheXpert), however, both lung masking and lung cropping show reduced race identification performance compared to the baseline (AUROC 0.566 and 0.593, respectively, vs. 0.623 for the baseline), indicating reduced racial encoding.
CLAHE yields a similar degree of racial encoding compared to the baseline.

We also evaluated average inter-racial differences in diagnostic model performance across all disease labels.
The average diagnostic AUROC differences on the internal test set are 0.0325 (Baseline), 0.0282 (CLAHE), 0.0352 (Cropping), and 0.0316 (Masking), and 0.0781 (Baseline), 0.0706 (CLAHE), 0.0678 (Cropping), and 0.0792 (Masking) on the external test set.

\section{Discussion}
In this study, we evaluated whether simple preprocessing can improve the generalization and robustness of chest X-ray classifiers while mitigating racial bias. We compared four pipelines -- a baseline approach with no additional preprocessing, lung masking, lung cropping, and CLAHE -- across internal and external evaluation datasets. Our findings indicate that simple lung cropping can yield reduced racial encoding (and, thus, reduced risk for racial biases) while maintaining high diagnostic performance. Lung masking, while reducing racial encoding, resulted in a notable drop in external diagnostic performance as also reported by Sourget et al.~\cite{Sourget2025}, while CLAHE had no notable effect on either racial encoding or diagnostic performance.
Our study contributes to a growing body of evidence showing that there are no inherent fairness-accuracy trade-offs but that model fairness and diagnostic performance can be improved in tandem~\cite{Petersen2023a}.

For correctly interpreting our results, it is notable that our baseline is strong and implements various best practices emerging from prior research, including a limitation to frontal CXR recordings~\cite{lotter2024acquisition}, preferential sampling of images with the maximum number of diseases labeled to reduce the risk of label bias~\cite{Weng2023} and data augmentation~\cite{wang2024drop}, all of which have been shown to reduce the risk of bias.
It is conceivable that the impact of preprocessing methods such as the ones investigated here might differ in a more basic setting with higher risk of bias.

Future research should systematically study the effect of CLAHE hyperparameters -- particularly the clip limit and grid tile size -- on model generalization, assessing whether principled tuning can yield gains in model fairness and robustness. In addition, advanced masking strategies, such as inpainting or gated or partial convolutions that natively handle missing data~\cite{liu2018image}, could reduce information loss and boundary artifacts, potentially resulting in better performance of the lung masking approach.

\printbibliography

\end{document}